%% file: main.tex
\documentclass[10pt,twocolumn,letterpaper]{article}
\pdfoutput=1

\usepackage{cvpr}
\usepackage{caption}
\usepackage{times}
\usepackage{epsfig}
\usepackage{graphicx}
\usepackage{amsmath}
\usepackage{amssymb}
\usepackage{xspace}
\usepackage{algorithm}
\usepackage{algorithmic}
\usepackage{multirow}
\usepackage{subcaption}

\usepackage[usenames, dvipsnames]{color} 
\definecolor{purple}{rgb}{0.5, 0.0, 0.5}
\definecolor{orange}{rgb}{1, 0.65, 0}
\definecolor{lightgreen}{rgb}{0.68, 1, 0.18}
\definecolor{darkgreen}{rgb}{0.09, 0.32, 0.24}
\definecolor{darkred}{rgb}{0.6, 0, 0}
\definecolor{brown}{rgb}{0.64, 0.16, 0.16}
\iffalse 
  \newcommand{\holger}[1]{\noindent}
  \newcommand{\oscar}[1]{\noindent}
  \newcommand{\alex}[1]{\noindent}
  \newcommand{\sourabh}[1]{\noindent}
  \newcommand{\done}[1]{\noindent}
  \newcommand{\todo}[1]{\noindent}
\else
  \newcommand{\holger}[1]{\textcolor{blue}{\bf [HC: #1]}}
  \newcommand{\oscar}[1]{\textcolor{orange}{\bf [OB: #1]}}
  \newcommand{\alex}[1]{\textcolor{purple}{\bf [AL: #1]}}
  \newcommand{\sourabh}[1]{\textcolor{brown}{\bf [SV: #1]}}
  \newcommand{\done}[1]{\textcolor{darkgreen}{\bf [Done: #1]}}
  \newcommand{\todo}[1]{\textcolor{red}{\bf [Todo: #1]}}
\fi
\usepackage[normalem]{ulem} 

\newcommand{\figref}[1]{Figure \ref{#1}}
\newcommand{\tableref}[1]{Table \ref{#1}}

\newcommand{\squeeze}{\vspace{-0.5mm}}

\usepackage[pagebackref=true,breaklinks=true,letterpaper=true,colorlinks,bookmarks=false]{hyperref}

\cvprfinalcopy 

\def\cvprPaperID{6374} 

\begin{document}

\title{\textbf{Real-Time 3D Object Detection Using InnovizOne LiDAR and Low-Power Hailo-8 AI Accelerator}}

\author{
\begin{tabular}[t]{ccc}
Itay Krispin-Avraham$^{1}$ & Roy Orfaig$^{2}$ & Ben-Zion Bobrovsky$^{2}$ \\
{\tt\footnotesize itaykrispin@mail.tau.ac.il} & {\tt\footnotesize royorfaig@tauex.tau.ac.il} & {\tt\footnotesize bobrov@tauex.tau.ac.il}
\end{tabular}
\\
{\small $^{1}$Faculty of Exact Sciences, Tel-Aviv University} \\
{\small $^{2}$School of Electrical Engineering, Tel-Aviv University}
}
\maketitle

\begin{abstract}
Object detection is a significant field in autonomous driving. Popular sensors for this task include cameras and LiDAR sensors. LiDAR sensors offer several advantages, such as insensitivity to light changes, like in a dark setting (\figref{fig:innoviz_pointcloud2}) and the ability to provide 3D information in the form of point clouds, which include the ranges of objects. However, 3D detection methods, such as \textit{PointPillars \cite{lang2019pointpillarsfastencodersobject}}\textit{, typically require high-power hardware. Additionally, most common spinning LiDARs are sparse and may not achieve the desired quality of object detection in front of the car. In this paper, we present the feasibility of performing real-time 3D object detection of cars using 3D point clouds from a LiDAR sensor, processed and deployed on a low-power Hailo-8 AI }\textit{accelerator} \cite{Hailo-8}. The LiDAR sensor used in this study is the\textit{ InnovizOne }\textit{sensor \cite{InnovizOne}, which captures objects in higher quality compared to spinning LiDAR techniques, especially for distant objects. We successfully achieved real-time inference at a rate of approximately 5Hz with a high accuracy of 0.91\% F1 score, with only  -0.2\% degradation compared to running the same model on an NVIDIA GeForce RTX 2080 Ti \cite{nvidia2080}. This work demonstrates that effective real-time 3D object detection can be achieved on low-cost, low-power hardware, representing a significant step towards more accessible autonomous driving technologies.  The source code and the pre-trained models are available at 
\url{https://github.com/AIROTAU/PointPillarsHailoInnoviz/tree/main}}.
\end{abstract}
\input{sections/introduction.tex}

\input{sections/methodology.tex}

\input{sections/integration.tex}

\input{sections/results.tex}

\input{sections/discussion.tex}

\input{sections/conclusion.tex}

{\small
\bibliographystyle{ieee}
\bibliography{main.bib}
}

\end{document}

%% file: sections/introduction.tex
\squeeze
\section{Introduction} \label{sec:intro}
\squeeze

In the field of autonomous driving, 3D object detection is crucial for understanding the environment and making informed navigation and safety decisions. Traditional approaches often rely on high-power, expensive hardware to achieve real-time performance, posing challenges for scalability and cost-effectiveness. This study investigates the use of the InnovizOne LiDAR sensor in conjunction with the Hailo AI Accelerator, a low-power alternative, to perform real-time 3D object detection.

InnovizOne LiDAR provides high-resolution 3D point cloud data, essential for accurately detecting and classifying objects in various environments. The Hailo-8 AI chip (\figref{fig:Hailo8}) , designed for edge devices, offers a cost-effective and energy-efficient solution for deploying deep learning models. By leveraging the OpenPCDet framework \cite{OpenPCDet}, we adapted state-of-the-art detector, PointPillars, to work with InnovizOne LiDAR data and optimized them for real-time inference on the Hailo chip.

%% file: sections/methodology.tex
\squeeze
\section{\textbf{Methodology}} \label{sec:methodology}
\squeeze

PointPillars accepts point clouds as input and estimates oriented 3D boxes for objects such as cars, pedestrians and cyclists.
It consists of three main stages:
(1) A feature encoder network that converts a point cloud to a sparse pseudo-image;
(2) a 2D convolutional backbone to process the pseudo-image into a high-level representation; and
(3) a detection head that detects and regresses 3D bounding boxes.

\subsection{\textbf{Data Collection}}
\begin{figure*}[ht]
  \centering
  \begin{minipage}[t]{0.49\textwidth}
    \centering
    \includegraphics[width=\textwidth]{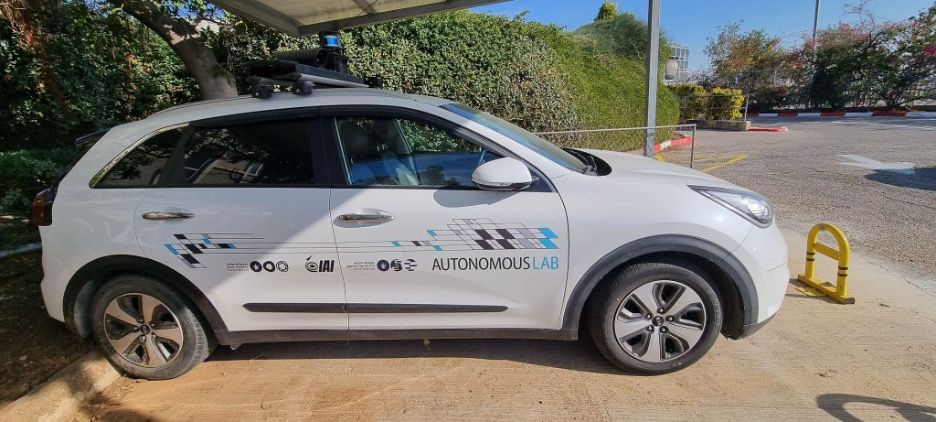}
    \label{fig:lab_vehicle}
  \end{minipage}
  \hfill
  \begin{minipage}[t]{0.49\textwidth}
    \centering
    \includegraphics[width=\textwidth]{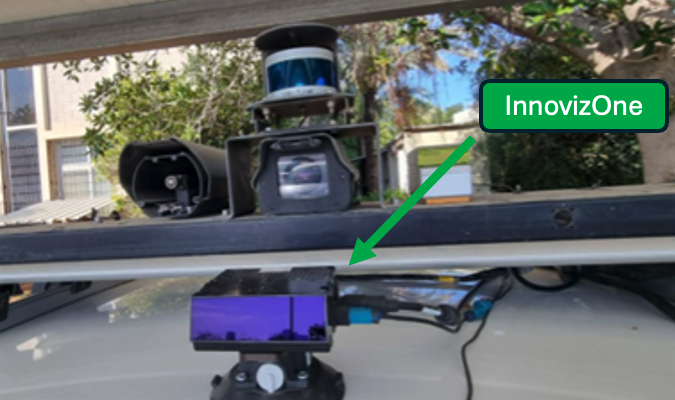}
    \label{fig:InnovizOne_Vehicle}
  \end{minipage}
  \caption{Left image: The lab vehicle equipped with a multi-sensor kit, including the InnovizOne LiDAR mounted at the front of the roof. Right image: A close-up of the InnovizOne LiDAR mounted on the vehicle’s roof}
  \label{fig:Vehicle_and_Sensor}
\end{figure*}

\begin{figure}[H]
    \centering
    \includegraphics[width=1\linewidth]{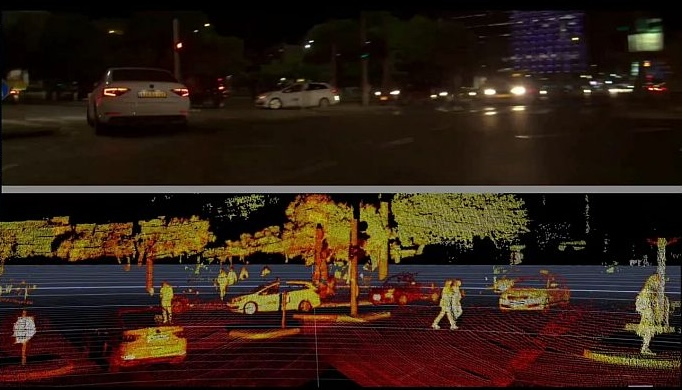}
    \caption{The image shows a traditional camera view at night, where pedestrians are difficult to spot due to poor lighting. The bottom image, generated using Innoviz LiDAR data, reveals a much clearer and detailed scene, detecting pedestrians and vehicles even in complete darkness. This demonstrates how LiDAR technology excels in low-light environments, offering critical advantages for autonomous systems. \cite{Innoviz_pointcloud_example}}
        \label{innoviz_pointcloud2}
    \label{fig:innoviz_pointcloud2}
\end{figure}

Our autonomous lab vehicle (\figref{fig:Vehicle_and_Sensor}), equipped with an InnovizOne LiDAR, was used by Yasmin Tsiprun in her work \cite{Dataset_creation} to record data in diverse environments, including both static and dynamic scenes across Tel-Aviv University campus and its surrounding roads. 
This type of LiDAR captures high-resolution 3D point clouds, providing detailed information about the area in front of the car. Unlike cameras, which can struggle in low light, LiDAR offers significant advantages by delivering detailed information about the surroundings in high-resolution 3D 
(\figref{fig:innoviz_pointcloud2}).

\squeeze
\subsection{\textbf{Data Annotation}}
\squeeze
The raw LiDAR data was divided into segments, with each frame labeled to identify cars. The labeled data was converted into a format compatible with the OpenPCDet framework, specifically preparing it for the PointPillars detector model. For each object, the annotated data include the 3D center of the objects (cx, cy and cz), the orientation of the objects in relation to the LiDAR (theta) and a 3D bounding box size (width, height and depth). We chose to utilize this dataset in our project.

\squeeze
\subsection{\textbf{Real-Time Inference on Hailo AI Accelerator }}
\squeeze
To achieve real-time performance, we integrated the trained detector with the Hailo AI accelerator. This involved several steps and optimizations to ensure that the model performed efficiently and accurately on the device, given its computational constraints compared to more powerful but costly alternatives like NVIDIA Jetson \cite{NVIDIA_Jetson}.

The Hailo-8 AI chip is designed for efficient edge computing, offering low power consumption, which makes it suitable for deployment in resource-constrained environments, and high computational efficiency, enabling real-time processing of high-resolution sensor data. 
\begin{figure}[H]
    \centering
    \includegraphics[width=1\linewidth]{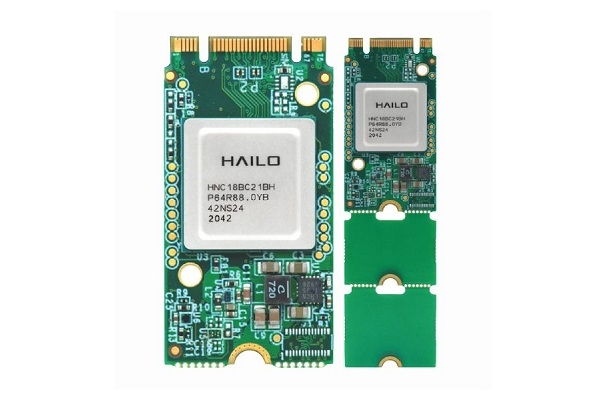}
    \caption{Hailo-8 AI Accelerator.}
    \label{fig:Hailo8}
\end{figure}

\squeeze
\section{Pipeline Overview} 
\squeeze
To integrate the PointPillars model with the Hailo-8 AI accelerator, we utilized Hailo's proof-of-concept (POC) \cite{Hailo_POC}, which demonstrated the offloading of computationally intensive 2D-convolutional layers of a 3D object detection network operating on point clouds from the KITTI dataset \cite{liao2022kitti360noveldatasetbenchmarks}. We adapted this POC to process data captured by our InnovizOne LiDAR. The pipeline for this POC involves the following steps:

\subsection{Data Preparation and Preprocessing:}

\begin{itemize}
    \item \textbf{Conversion to Compatible Formats:} The raw InnovizOne LiDAR data was converted into a format compatible with the PointPillars model and the Hailo hardware. This involved segmenting the data into frames and labeling objects within each frame for the model training phase.
    \item \textbf{Normalization and Augmentation:} The PCDet framework performs data normalization and augmentation in order to improve model robustness. This included transformations such as random flips and rotations to simulate different environmental conditions.
\end{itemize}

\subsection{Model Adaptation for Hailo:}

To leverage the Hailo hardware for accelerating the 2D-convolutional parts of the PointPillars network, the model architecture and execution flow was adapted. This involves several critical steps: exporting the PyTorch model to ONNX \cite{ONNX}, translating the ONNX model to a Hailo-compatible format, and creating a new PyTorch module that integrates the Hailo-inferred operations. The process included the following kep steps:

\subsubsection{\textbf{Extracting the 2D Backbone and Dense Head:}
}
\begin{figure*}[ht]
    \centering
    \includegraphics[width=0.75\linewidth]{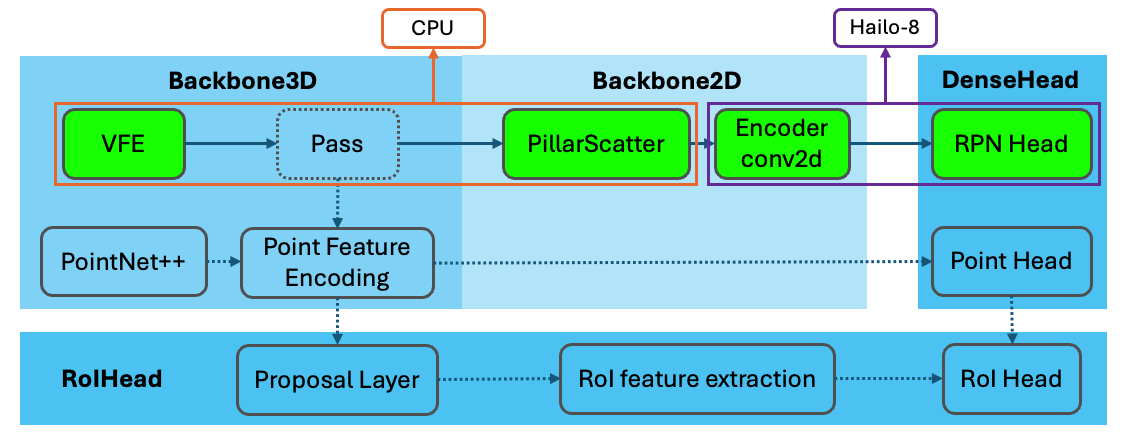}
    \caption{The diagram shows in orange the part of the PointPillars model architecture that is processed by the CPU and in purple the part that is being offloaded to the Hailo-8 accelerator.}
    \label{fig:architecture}
\end{figure*}

The 2D convolutional layers and the detection head were isolated from the PointPillars network. These components are responsible for most of the computational load and are well-suited for offloading to the Hailo hardware (\figref{fig:architecture}). A new PyTorch module was created to encapsulate these components. This module takes the spatial features as input and produces intermediate features, classification predictions, and bounding box predictions.

\subsubsection{\textbf{Exporting to ONNX:}
}
The newly created module was exported to the ONNX format. This format serves as an intermediary representation that can be parsed and optimized by Hailo tools. The ONNX model was simplified using onnxsim to ensure compatibility and efficiency in subsequent steps.

\subsubsection{\textbf{Translating to Hailo Internal Representation:}
}
Using the Hailo SDK, the simplified ONNX model was translated into Hailo's internal format (HAR). This process involves parsing the ONNX model and mapping its operations to Hailo's hardware capabilities. The resulting HAR file encapsulates the 2D convolutional layers and the detection head, ready for execution on Hailo hardware.

\subsubsection{\textbf{Creating a PyTorch Module for Hailo Execution:}}

A new PyTorch module was defined to replace the original 2D backbone and dense head in the PointPillars network. This module integrates with Hailo's hardware to perform the 2D convolution and detection head operations. It handles the conversion of data formats and interfaces with the Hailo SDK. This ensures that the spatial features are processed by the Hailo hardware and the results are seamlessly integrated back into the PyTorch model flow.

The system architecture was designed to maximize the efficiency of real-time 3D object detection while maintaining low power consumption. Key components included:

\begin{itemize}
    \item \textbf{Data Acquisition Module:} Captures LiDAR data and converts it into a format suitable for the processing pipeline.
    \item \textbf{Preprocessing Module:} Applies necessary transformations and augmentations to the raw data, ensuring it is ready for inference.
    \item \textbf{Inference Engine:} Runs the optimized PointPillars model on the Hailo-8 AI chip, performing real-time 3D object detection.
    \item \textbf{Post-Processing Module:} Refines the model output, filtering and merging detections to provide accurate and reliable results.
\end{itemize}

%% file: sections/integration.tex
\squeeze
\section{\textbf{Integration and Testing}} \label{sec:integration}
\squeeze

\textbf{Overview:} After adapting the model for Hailo, we integrated it into the overall inference pipeline and performed testing. This ensured that the adapted model produced consistent and accurate results and leveraged the Hailo hardware effectively. The process included the following kep steps:

\subsection{\textbf{Quantization and Optimization:}}

\begin{itemize}
    \item To further optimize the model for Hailo hardware, a quantization process was performed. This process involved creating a calibration dataset from the spatial features input to the 2D backbone.
    \item Using the Hailo SDK, the model based on the calibration dataset was optimized. This step ensured that the model was numerically efficient and ready for execution on Hailo hardware.
    \item The optimized model was saved as a quantized HAR (q-HAR) file.
\end{itemize}

\subsection{\textbf{Compiling for Hailo Hardware:}}

\begin{itemize}
    \item The quantized model was then compiled for execution on Hailo hardware. This involved generating a hardware executable file (HEF) that encapsulates the optimized model.
    \item The compilation process included creating an allocation plan and optimizing the resource utilization on the Hailo hardware. The resulting HEF file was ready for deployment.
\end{itemize}

\subsection{\textbf{End-to-End Integration:}}

\begin{itemize}
    \item The Hailo hardware execution was integrated into the overall inference pipeline using HailoRT's asynchronous send/receive functionality. This allowed for efficient and pipelined processing.
    \item Two new PyTorch modules were defined to handle the operations before and after the Hailo-mapped parts. These modules encapsulated the preprocessing and postprocessing steps, ensuring a seamless flow of data.
    \item A multiprocessing setup was implemented to manage the data transfer between PyTorch and Hailo hardware. Separate processes handled the sending and receiving of data, enabling efficient and parallel execution.
\end{itemize}

\subsection{\textbf{Final Testing:}}
\begin{itemize}
    \item Extensive testing was performed to verify the accuracy and performance of the integrated model. We compared the results with the original model to ensure consistency.
    \item The end-to-end inference pipeline was validated to ensure it met the required performance metrics and utilized the Hailo hardware effectively.
\end{itemize}

%% file: sections/results.tex
\begin{table*}[ht]
\small
\centering
\begin{tabular}{|c|c|c|c|c|c|} \hline  

Model & F1 Score & Recall & Precision & AP \\ \hline  

PVRCNN & \textbf{0.96} & \textbf{0.97} & 0.96 & \textbf{0.97} \\ \hline  

PointPillars & 0.92 & 0.87 & \textbf{0.97} & 0.87 \\ \hline  

PointPillars+Hailo (ours) & 0.91 & 0.87 & 0.96 & 0.85 \\ \hline 
\end{tabular}
\caption{The result metrics \cite{powers2020evaluationprecisionrecallfmeasure} were evaluated using an IoU threshold and a confidence threshold of 0.3. We evaluated the results for two models (PVRCNN and PointPillars) and for the new pipeline created with the offloaded computation to the Hailo chip based on the PointPillars model.} 
\label{table:res_detection}
\end{table*}

\squeeze
\section{\textbf{Results}}\label{sec:results}
\squeeze
The optimized model on the Hailo chip achieved a processing rate of approximately 5 Hz, with detection accuracy comparable to running on more powerful hardware. This demonstrated the feasibility of deploying advanced 3D object detection models on low-power edge devices. In addition to the PointPillars model, we also trained the PV-RCNN model \cite{shi2021pvrcnnpointvoxelfeatureset}, which is a more complex and powerful 3D object detection architecture. PV-RCNN is known for its superior accuracy due to its multi-scale feature aggregation and region-based refinement. However, the model is computationally heavier, and we recognized early on that it would not be feasible to run PV-RCNN on the Hailo chip, given its resource constraints. Despite this, we conducted the comparison to evaluate the potential trade-offs between model complexity and performance. While PV-RCNN achieved higher accuracy on the same dataset, its processing speed was significantly slower, further justifying our choice to optimize the lighter PointPillars model for real-time edge inference on the Hailo platform.

The experimental results, summarized in \tableref{table:res_detection}, highlight the performance metrics for the PV-RCNN and PointPillars models. Evaluations were conducted using an IoU threshold and a confidence threshold of 0.3, with comparisons made between the models' performance with and without the Hailo-optimized pipeline. Notably, the PointPillars model, optimized for the Hailo chip, demonstrated competitive accuracy while achieving faster processing speeds. Detection metrics for PointPillars at these thresholds showed minimal variation between the standard and Hailo-optimized pipelines.

%% file: sections/discussion.tex
\squeeze
\section{\textbf{Discussion}} \label{sec:discussion}
\squeeze

 \begin{figure*}[ht]
  \centering
  \begin{minipage}[t]{0.49\textwidth}
    \centering
    \includegraphics[width=\textwidth]{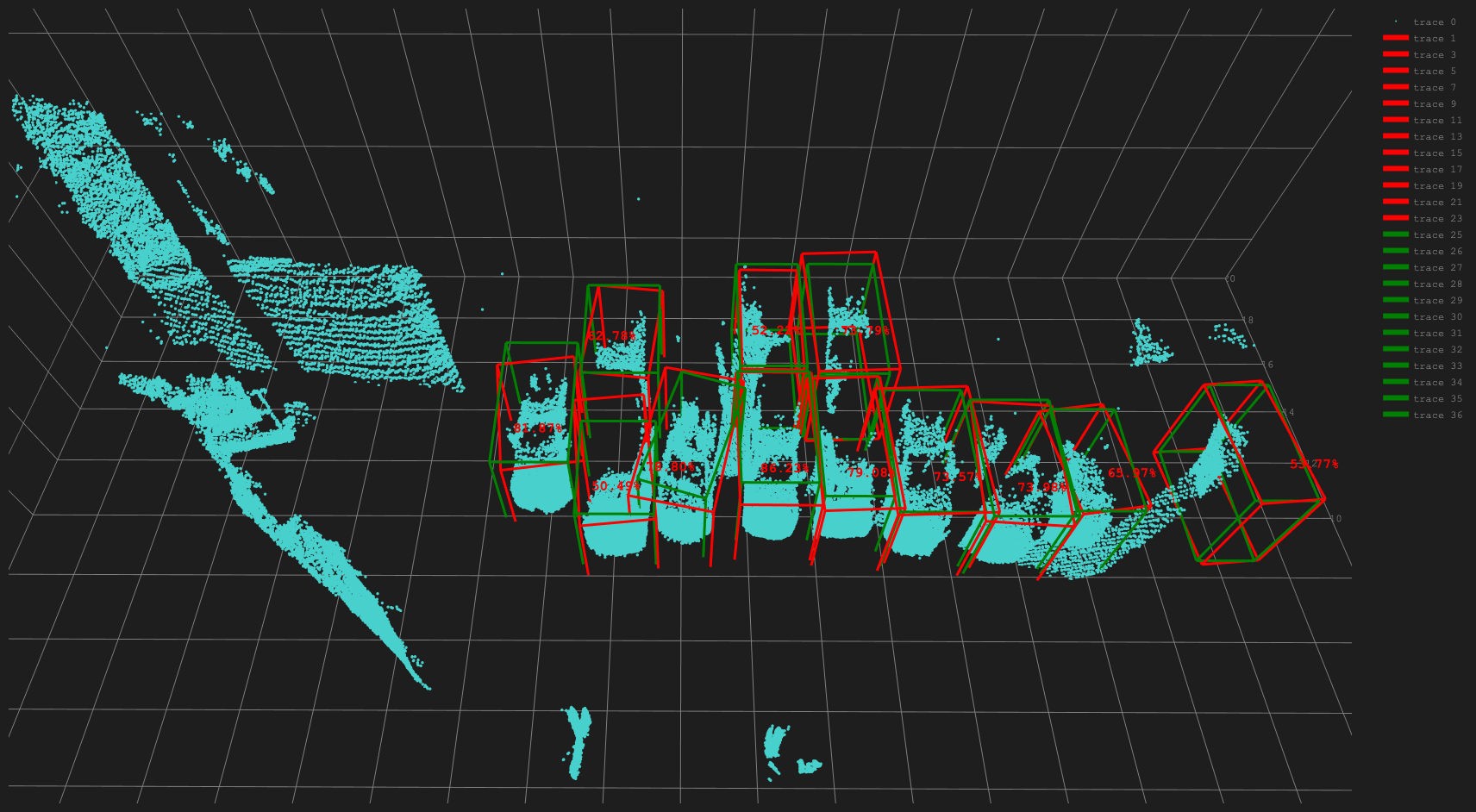}
    \label{fig:00001}
  \end{minipage}
  \hfill
  \begin{minipage}[t]{0.49\textwidth}
    \centering
    \includegraphics[width=\textwidth]{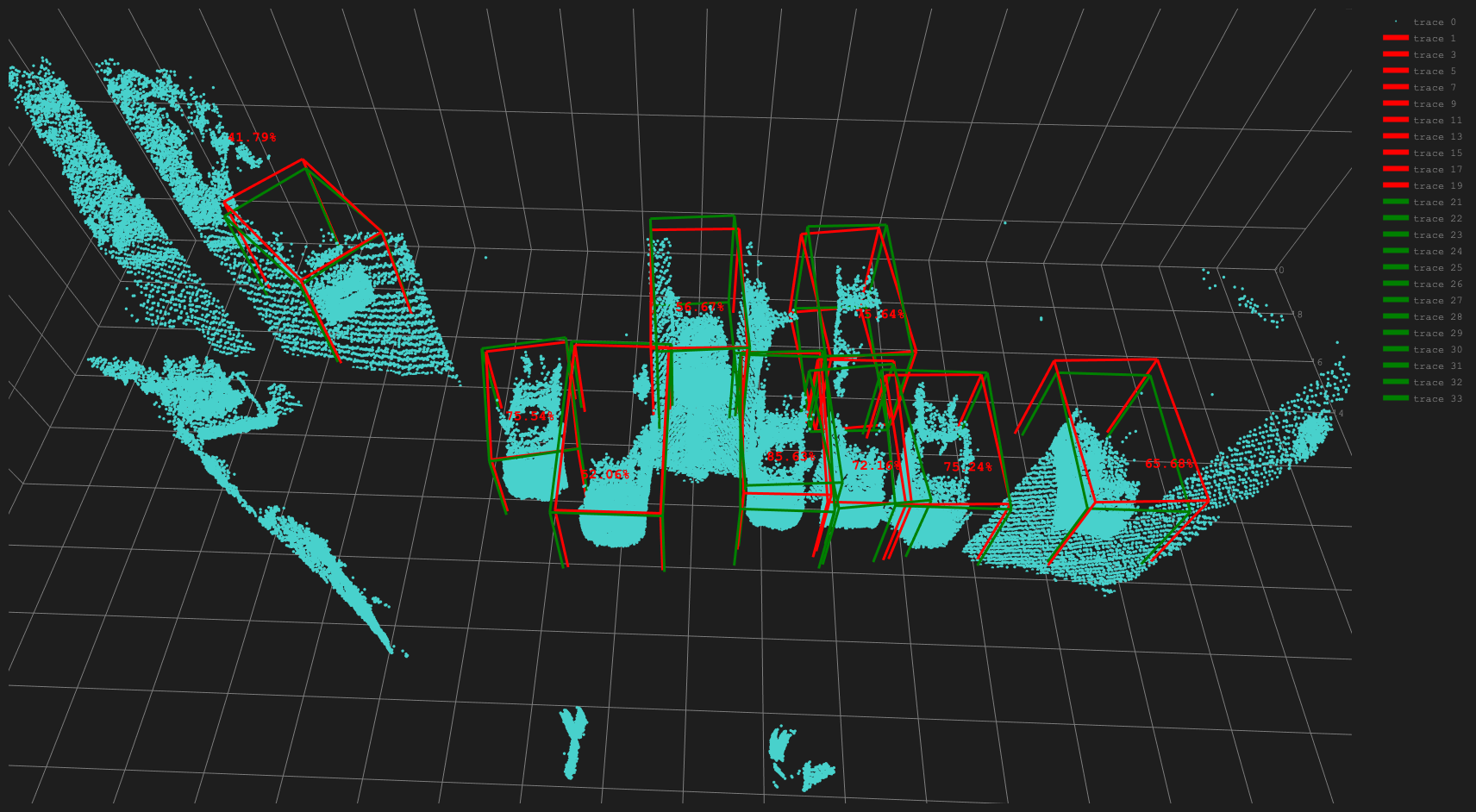}
    \label{fig:00101}
  \end{minipage}
  \vskip\baselineskip
  \begin{minipage}[t]{0.49\textwidth}
    \centering
    \includegraphics[width=\textwidth]{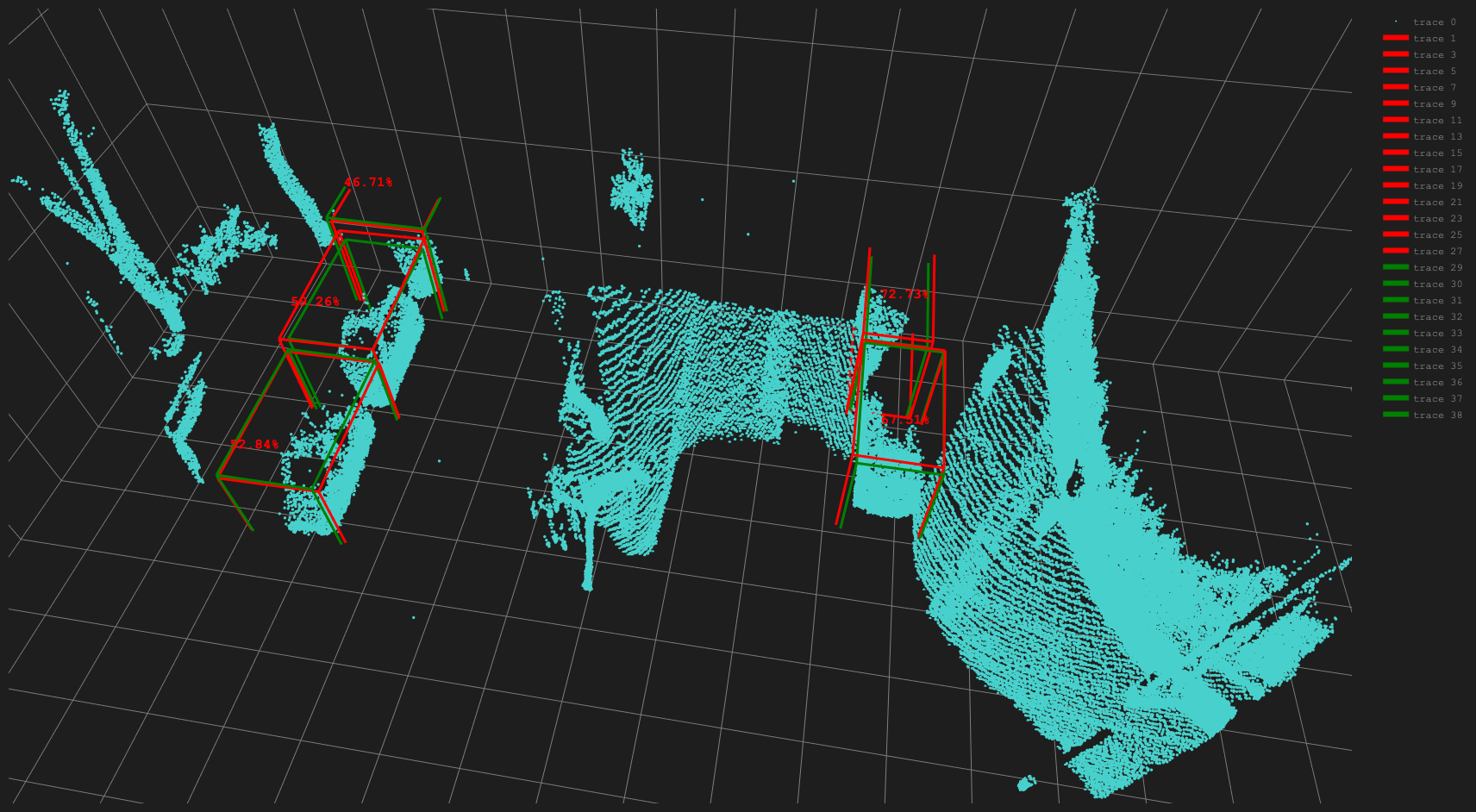}
    \label{fig:00151}
  \end{minipage}
  \hfill
  \begin{minipage}[t]{0.49\textwidth}
    \centering
    \includegraphics[width=\textwidth]{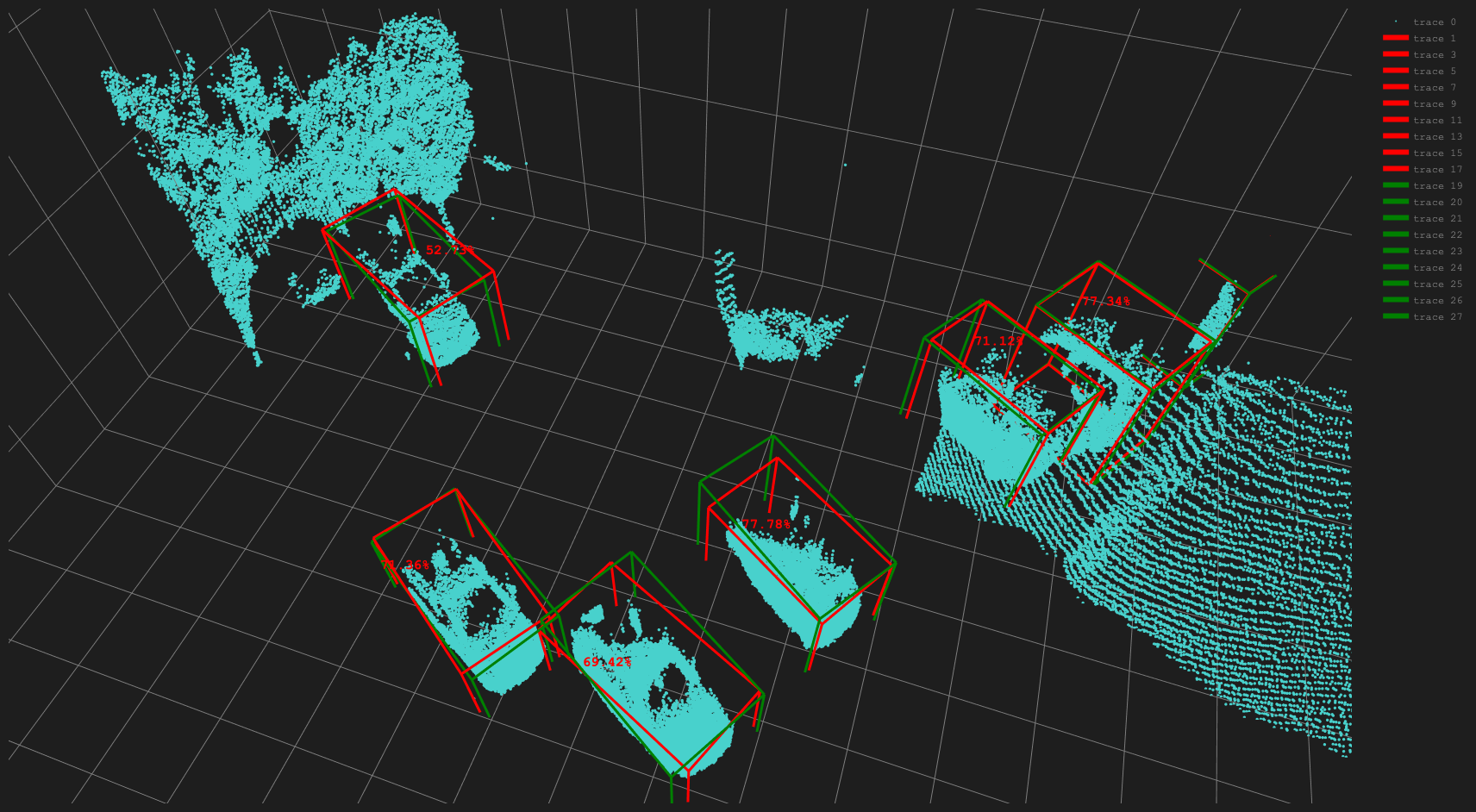}
    \label{fig:00298}
  \end{minipage}
  \caption{Above are several detection of cars generated by the PointPillars model using Hailo on point cloud data from Innoviz LiDAR. Red bounding boxes represent the detector results, while green bounding boxes indicate the ground truth.\\}
  \label{fig:collage}
\end{figure*}

The successful deployment of real-time 3D object detection on low-power hardware is a major step forward for the future of autonomous driving technologies. This advancement not only demonstrates the feasibility of using efficient, power-constrained hardware in complex perception tasks but also highlights the potential for broader adoption across various autonomous platforms. By addressing challenges related to scalability and cost, our approach enables more affordable access to advanced detection capabilities, which could benefit a wide range of applications, from commercial autonomous vehicles to agricultural and industrial automation.

In addition, our study identifies key areas for further development, including optimization techniques that could reduce processing latency and enhance detection accuracy. Expanding the applicability of our approach to integrate with various sensor types, such as other LiDAR models and radar, could improve system robustness across diverse driving conditions and environments. These ongoing improvements will be essential for advancing the reliability and versatility of real-time 3D object detection in autonomous systems

%% file: sections/conclusion.tex
\squeeze
\section{Conclusion} \label{sec:conclusion}
\squeeze
This study illustrates that high-performance real-time 3D object detection is achievable using cost-effective, low-power hardware. By leveraging the capabilities of the InnovizOne LiDAR sensor in combination with the Hailo AI chip, and through careful optimization of the PointPillars model architecture, we achieved a processing rate of 5Hz with substantial accuracy. This setup strikes a balance between performance and energy efficiency, showcasing the feasibility of deploying advanced perception systems without the need for expensive, high-power GPUs traditionally associated with autonomous vehicle technologies.

The results underscore the potential for more accessible and scalable autonomous driving solutions. This approach can be particularly impactful in applications where both cost and power consumption are limiting factors, such as compact or lightweight robotic systems, autonomous delivery vehicles, and agricultural or industrial automation. Moreover, it opens opportunities to integrate high-resolution 3D perception in environments with constrained resources, making autonomous technology more adaptable and affordable across a wider range of sectors. Future work could extend these optimizations to support additional sensor types and processing frameworks, further broadening the scope and accessibility of autonomous systems.

\squeeze
\paragraph{\textbf{Acknowledgment}} \label{sec:conclusion} \mbox{} \\
\squeeze

We thank both Innoviz and Hailo for their generous donations of the LiDAR and the AI accelerators. We also thank Yasmin Tsiprun for her work on collecting and creating the point-cloud dataset and allowing us to use it in this project. We would also like to thanks Roi Raich for all the work and support he provided us along the way, including in the creation of the dataset.